\documentclass[10pt,twocolumn,letterpaper]{article}

\usepackage{cvpr}              % To produce the CAMERA-READY version
\usepackage{graphicx}
\usepackage{amsmath}
\usepackage{amssymb}
\usepackage{booktabs}
\usepackage{lipsum} 
\usepackage{multirow}
\usepackage[accsupp]{axessibility}  
\usepackage[pagebackref,breaklinks,colorlinks]{hyperref}

\usepackage[capitalize]{cleveref}
\crefname{section}{Sec.}{Secs.}
\Crefname{section}{Section}{Sections}
\Crefname{table}{Table}{Tables}
\crefname{table}{Tab.}{Tabs.}

\def\cvprPaperID{1856}
\def\confName{CVPR}
\def\confYear{2022}

\begin{document}
\newcommand{\qdet}{QueryDet}
\newcommand{\csqFull}{Cascade Sparse Query (CSQ)}
\newcommand{\csqTerm}{Cascade Sparse Query}
\newcommand{\csqAbrr}{CSQ}

\title{QueryDet: Cascaded Sparse Query for Accelerating High-Resolution Small Object Detection}

\author{Chenhongyi Yang\thanks{Work done when working as a full-time research intern at TuSimple.}\\
University of Edinburgh\\
{\tt\small chenhongyi.yang@ed.ac.uk}
\and
Zehao Huang\\
TuSimple\\
{\tt\small zehaohuang18@gmail.com}
\and
Naiyan Wang\\
TuSimple\\
{\tt\small winsty@gmail.com}
}

% Documents
\maketitle

\begin{abstract}
   While general object detection with deep learning has achieved great success in the past few years, the performance and efficiency of detecting small objects are far from satisfactory. The most common and effective way to promote small object detection is to use high-resolution images or feature maps. However, both approaches induce costly computation since the computational cost grows squarely as the size of images and features increases. To get the best of two worlds, we propose \qdet~that uses a novel query mechanism to accelerate the inference speed of feature-pyramid based object detectors. The pipeline composes two steps: it first predicts the coarse locations of small objects on low-resolution features and then computes the accurate detection results using high-resolution features sparsely guided by those coarse positions. In this way, we can not only harvest the benefit of high-resolution feature maps but also avoid useless computation for the background area. On the popular COCO dataset, the proposed method improves the detection mAP by 1.0 and mAP-small by 2.0, and the high-resolution inference speed is improved to 3.0$\times$ on average. On VisDrone dataset, which contains more small objects, we create a new state-of-the-art while gaining a 2.3$\times$ high-resolution acceleration on average. Code is available at \url{https://github.com/ChenhongyiYang/QueryDet-PyTorch}.
   
\end{abstract}
\section{Introduction}\label{sec:intro}
With the recent advances of deep learning~\cite{he2016deep, xie2017aggregated}, visual object detection has achieved massive improvements in both performance and speed \cite{girshick2014rich, liu2016ssd, redmon2018yolov3, ren2015faster, lin2017feature, lin2017focal,carion2020end, wang2020scale}. It has become the foundation for widespread applications, such as autonomous driving and remote sensing. However, detecting small objects is still a challenging problem. There is a large performance gap between small and normal scale objects. Taking RetinaNet \cite{lin2017focal}, one of the state-of-the-art object detectors, as an example, it achieves 44.1 and 51.2 mAP on objects with medium and large sizes but only obtains 24.1 mAP on small objects on COCO \cite{lin2014microsoft} \textit{test-dev} set. Such degradation is mainly caused by three factors: 1) the features that highlight the small objects are extinguished because of the down-sampling operations in the backbone of Convolutional Neural Networks (CNN); hence the features of small objects are often contaminated by noise in the background; 2) the receptive field on low-resolution features may not match the size of small objects as pointed in ~\cite{li2019scale}; 3) localizing small objects is more difficult than large objects because a small perturbation of the bounding box may cause a significant disturbance in the Intersection over Union (IoU) metric.

\begin{figure}[!t]
  \includegraphics[width=\columnwidth]{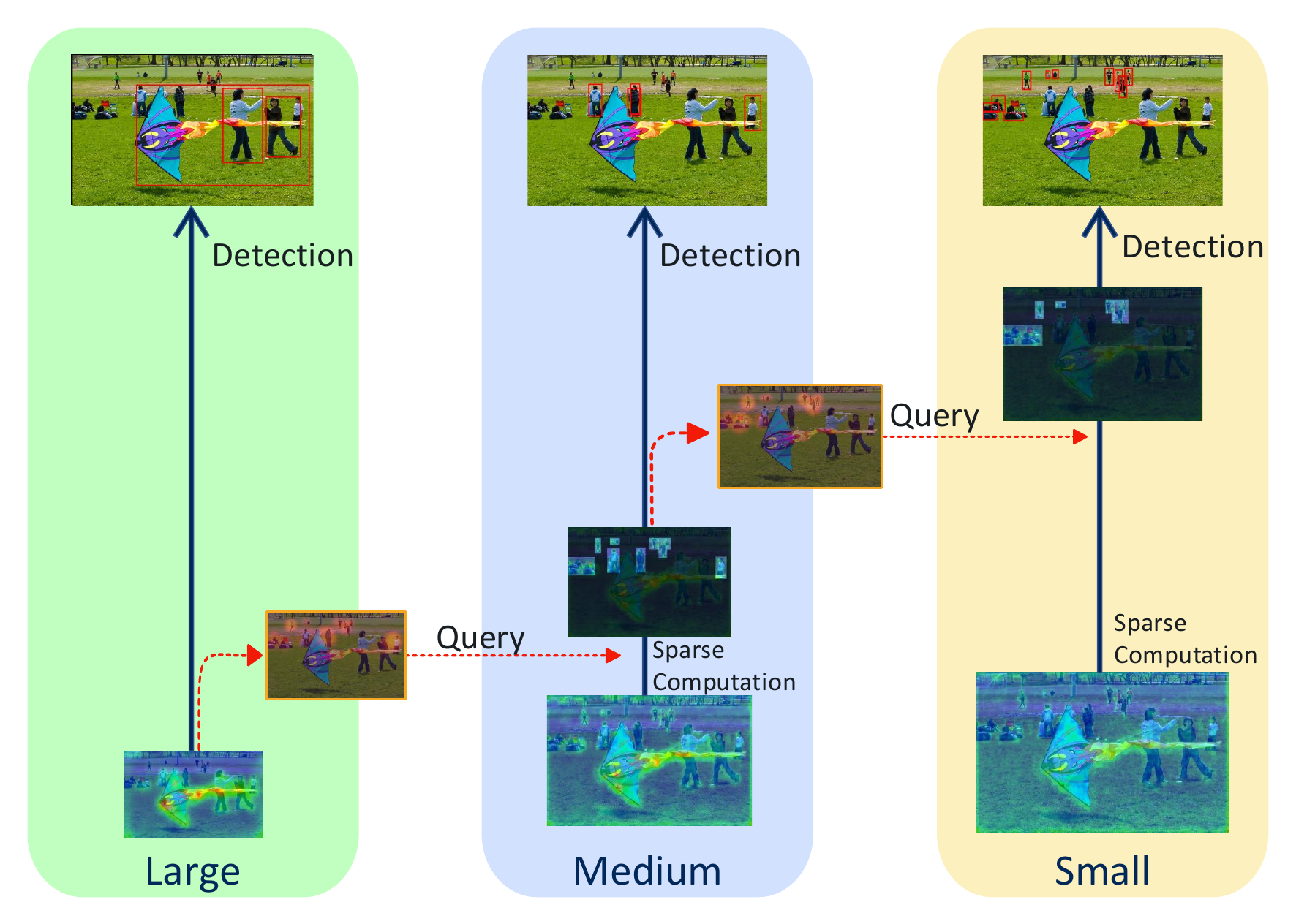}
  \caption{\qdet~achieves highly effective small object detection in high-resolution features. The locations (query keys) where small objects might exist are first predicted in the low-resolution features, and a sparse feature map (query values) is constructed using high-resolution features in those locations. Finally, a sparse detection head is used to output the detected boxes. This paradigm is applied in a cascaded manner, enabling fast and accurate small object detection.}
  \label{fig:fig1}
  \vspace{-0.3cm}
\end{figure}

Small object detection can be improved by scaling the size of input images or reducing the down-sampling rate of CNN to maintain high-resolution features, as they increase the effective resolution in the resulted feature map. However, merely increasing the resolution of feature maps can incur considerable computation costs. Several works~\cite{liu2016ssd, cai2016unified, lin2017feature} proposed to build a feature pyramid by reusing the multi-scale feature maps from different layers of a CNN to address this issue. Objects with various scales are handled on different levels: large objects tend to be detected on high-level features, while small objects are usually detected on low levels. The feature pyramid paradigm saves the computation cost of maintaining high-resolution feature maps from shallow to deep in the backbone. Nevertheless, the computation complexity of detection heads on low-level features is still enormous. For example, adding an extra pyramid level $P_2$ into RetinaNet will bring about 300\% more computation (FLOPs) and memory cost in the detection head; hence severely lowering down the inference speed from 13.6 FPS to 4.85 FPS on NVIDIA 2080Ti GPU.

In this paper, we propose a simple and effective method, \qdet, to save the detection head's computation while promoting the performance of small objects. The motivation comes from two key observations:
1) the computation on low-level features is highly redundant. In most cases, the spatial distribution of small objects is very sparse: they occupy only a few portions of the high-resolution feature maps; hence a large amount of computation is wasted. 2) The feature pyramids are highly structured. Though we cannot accurately detect the small objects in low-resolution feature maps, we can still infer their existence and rough locations with high confidence.

A natural idea to utilize these two observations is that we can only apply the detection head to small objects' spatial locations. This strategy requires locating the rough location of small objects at a low cost and sparse computation on the desired feature map.
In this work, we present \qdet~that is based on a novel query mechanism \csqFull, as illustrated in Fig.~\ref{fig:fig1}. We recursively predict the rough locations of small objects (queries) on lower resolution feature maps and use them to guide the computations in higher resolution feature maps. With the help of sparse convolution \cite{graham2017submanifold,yan2018second}, we significantly reduce the computation cost of detection heads on low-level features while keeping the detection accuracy for small objects. Note that our approach is designed to save the computation spatially, so it is compatible with other accelerating methods like light-weighted backbones~\cite{tan2020efficientdet}, model pruning~\cite{he2017channel}, model quantization~\cite{wei2018quantization}, and knowledge distillation~\cite{chen2017learning}.

We evaluate our \qdet~on the COCO detection benchmark \cite{lin2014microsoft} and a challenging dataset, VisDrone~\cite{zhu2018visdrone}, that contains a large amount of small objects. We show our method can significantly accelerate inference while improving the detection performance. In summary, we make two main contributions:
\begin{itemize}
	\item
	We propose \qdet, in which a simple and effective \csqFull~mechanism is designed. It can reduce the computation costs of all feature pyramid based object detectors. Our method can improve the detection performance for small objects by effectively utilizing high-resolution features while keeping fast inference speed.
	\item
	On COCO, \qdet~improves the RetinaNet baseline by 1.1 AP and 2.0 AP$_S$ by utilizing high-resolution features, and the high-resolution detection speed is improved by 3.0$\times$ on average when \csqAbrr~is adopted. On VisDrone, we advance the state-of-the-art results in terms of the detection mAP and enhance the high-resolution detectopm speed by 2.3$\times$ on average.
\end{itemize}

\section{Related Works}
\noindent \textbf{Object Detection.}
Deep Learning based object detection can be mainly divided into two streams: the two-stage detectors \cite{girshick2014rich, girshick2015fast, ren2015faster, lin2017feature, cai2018cascade} and the one-stage detectors \cite{huang2015densebox, redmon2016you, redmon2017yolo9000, redmon2018yolov3, liu2016ssd,zhang2020bridging} pioneered by YOLO. Generally speaking, two-stage methods tend to be more accurate than one-stage methods because they use the RoIAlign operation~\cite{he2017mask} to align an object's features explicitly. However, the performance gap between these two streams is narrowed recently. RetinaNet \cite{lin2017focal} is the first one-stage anchor-based detector that matches the performance of two-stage detectors. It uses feature pyramid network (FPN) \cite{lin2017feature} for multi-scale detections and proposes FocalLoss to handle the foreground-background imbalance problem in dense training. Recently, one-stage anchor-free detectors \cite{law2018cornernet, tian2019fcos, duan2019centernet, duan2019centernet, yang2019reppoints, kong2020foveabox} have attracted academic attentions because of their simplicity. In this paper, we implement our QueryDet based on RetinaNet and FCOS \cite{tian2019fcos} to show its effectiveness and generalization ability. \\

\noindent \textbf{Small Object Recognition.}
Small object recognition, like detection and segmentation, is a challenging computer vision task because of low-resolution features. To tackle this problem, a large amount of works have been proposed. These methods can be mainly categorized into four types: 1) increasing the resolution of input features \cite{liu2016ssd, cai2016unified, kong2016hypernet, fu2017dssd, lin2017feature, shrivastava2016beyond, li2017perceptual, wang2020deep}; 2) oversampling and strong data augmentation \cite{liu2016ssd, kisantal2019augmentation, zoph2019learning}; 3) incorporating context information \cite{chen2017deeplab, yu2015multi, chen2016r}, and 4) scale-aware training \cite{lin2017feature, singh2018analysis, singh2018sniper, li2019scale}. \\

\noindent \textbf{Spatial Redundancy.}
Several methods have used sparse computation to utilize the spatial redundancy of CNNs in different ways to save computation costs. PerforatedCNN~\cite{figurnov2016perforatedcnns} generates masks with different deterministic sampling methods. Dynamic Convolution~\cite{Verelst_2020_CVPR} uses a small gating network to predict pixel masks, and \cite{xie2020spatially} proposes a stochastic sampling and interpolation network. Both of them adopt Gumbel-Softmax \cite{jang2016categorical} and a sparsity loss for the training of sparse masks. On the other hand, the Spatially Adaptive Computation Time (SACT) \cite{figurnov2017spatially} predicts a halting score for each spatial position that is supervised by a proposed ponder cost and the task-specific loss function. SBNet \cite{SBNet} adopts an offline road map or a mask to filter out ignored region. Unlike these methods, our QueryDet focuses on objects' scale variation and simply adopts the provided ground-truth bounding box for supervision. Another stream of works adopts a two-stage framework: glance and focus for adaptive inference. \cite{wang2020glance} selects small regions from the original input image by reinforcement learning and processes these regions with a dynamic decision process. \cite{uzkent2020efficient} adopts a similar idea on object detection task. One similar work to our \qdet~is AutoFocus \cite{najibi2019autofocus}. AutoFocus first predicts and crop region of interest in coarse scales, then scaled to a larger resolution for final predictions. Compared with AutoFocus, our \qdet~is more efficient since the ``focus'' operation is conducted on feature pyramids other than image pyramids, which reduces the redundant computation in the backbone.

\begin{figure}[!t]
    \includegraphics[width=\columnwidth]{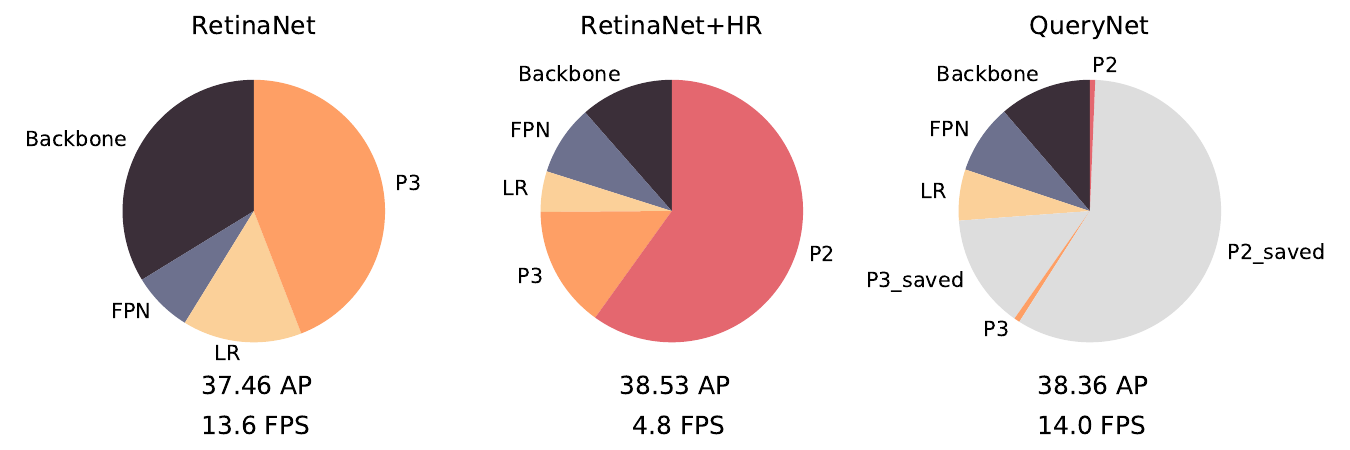}
    \caption{The FLOPs distribution of different module when using ResNet-50 backbone. In RetinaNet, the computational cost on the high-resolution $P_3$ accounts for 43\% of the total cost; when the higher-resolution $P_2$ is added, they together account for 74\% of the total cost. Our \qdet~can effectively reduce the computation on those features by 99\%, leading to a fast inference speed and keeping a high detection accuracy. Note LR stands for the low-resolution $P_4$ to $P_7$.}
    \label{fig:flops}
    \vspace{-0.2cm}
  \end{figure}

\section{Methods}
In this section, we describe our \qdet~for accurate and fast small object detection. We illustrate our approach based on RetinaNet \cite{lin2017focal}, a popular anchor-based dense detector. Note that our approach is not limited to RetinaNet, as it can be applied to any one-stage detectors and the region proposal network (RPN) in two-stage detectors with FPN. We will first revisit RetinaNet and analyze the computational cost distribution of different components. Then we will introduce how we use the proposed \csqTerm~to save computation costs during inference. Finally, the training details will be presented.

\subsection{Revisiting RetinaNet}\label{sec:method-1}
RetinaNet has two parts: a backbone network with FPN that outputs multi-scale feature maps and two detection heads for classification and regression. When the size of input image is $H\times W$, the sizes of FPN features are $\mathcal{P}=\{P_l\in \mathbb{R}^{H' \times W' \times C} \}$. Here $l$ indicates the pyramid level and $(H',W')$ is usually equals to $(\lfloor \frac{H}{2^l} \rfloor, \lfloor \frac{W}{2^l} \rfloor)$ in a typical FPN implementation. The detection heads consist of four $3 \times 3$ convolution layers, followed by an extra $3 \times 3$ convolution layer for final prediction. For parameter efficiency, different feature levels share the same detection heads (parameters). However, the computation costs are highly imbalanced across different layers: the FLOPs of detection heads from $P_7$ to $P_3$ increases in quadratic order by the scaling of feature resolutions. As shown in Figure~\ref{fig:flops}, the $P_3$ head occupies nearly half FLOPs while the cost of low-resolution features $P_4$ to $P_7$ only accounts for 15\%. Thus, if we want to extend the FPN to $P_2$ for better small object performance, the cost is unaffordable: high-resolution $P_2$ and $P_3$ will occupy 75\% of the overall cost. In the following, we describe how our \qdet~reduce the computation on high-resolution features and promote the inference speed of RetinaNet, even with an extra high-resolution $P_2$.

\subsection{Accelerating Inference by Sparse Query}\label{sec:method-infer}
\begin{figure*}[t]
\begin{center}
    \includegraphics[width=\linewidth]{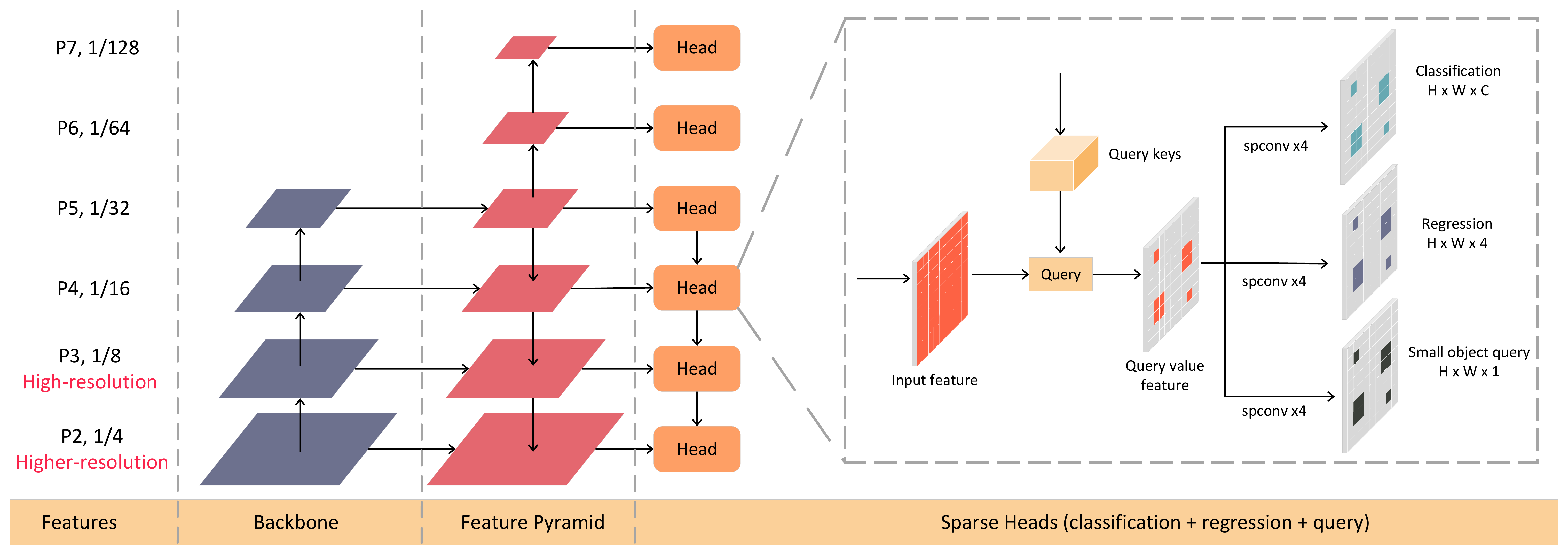}
\caption{The whole pipeline of the proposed \qdet. The image is fed into the backbone and Feature Pyramid Network (FPN) to produce a series of feature maps of different resolutions. Beginning from the query start layer ($P_5$ in this image), each layer receives a set of key positions from previous layer and a query operation is applied to generate the sparse value feature map. Then the sparse detection head and the sparse query head predict the detected boxes of the corresponding scales and key positions for the next layer.}
\label{fig:pipeline}
\end{center}
\vspace{-0.6cm}
\end{figure*}

In the design of modern FPN based detectors, small objects tend to be detected from high-resolution low-level feature maps. However, as the small objects are usually sparsely populated in space, the dense computation paradigm on high-resolution feature maps is highly inefficient. Inspired by this observation, we propose a coarse-to-fine approach to reduce the computation cost of low-level pyramids: First, the rough locations of small objects are predicted on coarse feature maps, and then the corresponding locations on fine feature maps are intensively computed. This process can be viewed as a query process: the rough locations are query keys, and the high-resolution features used to detect small objects are query values; thus we call our approach \textbf{\qdet}. The whole pipeline of our method is presented in Figure~\ref{fig:pipeline}.

To predict the coarse locations of small objects, we add a query head that is parallel to the classification and regression heads. The query head receives feature map $P_l$ with stride $2^l$ as input, and output a heatmap $V_l \in \mathbb{R}^{H' \times W'}$ with $V_l^{i,j}$ indicating the probability of that the grid $(i, j)$ contains a small object. During training, we define small objects on each level as objects whose scale is smaller than a pre-defined threshold $s_l$. Here we set $s_l$ to the minimum anchor scale on $P_l$ for simplicity, and for anchor-free detectors it is set to the minimum regression range on $P_l$.  For a small object $o$, we encode the target map for Query Head by computing distance between its center location $(x_o, y_o)$ and every location on the feature map, and set locations whose distance is smaller than $s_l$ to 1, otherwise 0. Then the Query Head is trained using FocalLoss~\cite{lin2017focal}. During inference, we choose the locations whose predicted scores are larger than a threshold $\sigma$ as queries. Then $q_l^o$ will be mapped to its four nearest neighbors on $P_{l-1}$ as key positions $\{k_{l-1}^o\}$:
\begin{align}
    \{k_{l-1}^o\} = \{(2x_l^o+i,2y_l^o+j), \forall i,j\in \{0,1\} \}.
\end{align}
All $\{k_{l-1}^o\}$ on $P_{l-1}$ are collected to form the key position set $\{k_{l-1}\}$. Then the three heads will only process those positions to detect objects and compute next level's queries. Specifically, we extract features from $P_{l-1}$ using $\{k_{l-1}\}$ as indices to construct a sparse tensor $P_{l-1}^v$ that we call value features. Then the sparse convolution (spconv)~\cite{graham2017submanifold} kernels are built using weights of the 4-\emph{conv} dense heads to compute results on layer $l-1$. 

To maximize the inference speed, we apply the queries in a cascade manner. In particular, the queries for $P_{l-2}$ would only be generated from $\{k_{l-1}\}$. We name this paradigm as \textbf{\csqFull}~as illustrated in Figure~\ref{fig:fig1}. The benefit of our \csqAbrr~is that we can avoid generating the queries $\{q_l\}$ from a single $P_{l}$, which leads to exponentially increasing size of corresponding key position ${k_l}$ during query mapping as $l$ decreases.

\subsection{Training}\label{sec:method-2}
We keep the training of classification and regression heads as same as in the original RetinaNet~\cite{lin2017focal}. For the query head, we train it using FocalLoss~\cite{lin2017focal} with the generated binary target map: Let the ground-truth bounding box of a small object $o$ on $P_l$ be $b_l^o=(x_l^o,y_l^o,w_l^o,h_l^o)$. We first compute the minimum distance map $D_l$ between each feature position $(x,y)$ on $P_l$ and all the small ground-truth centers $\{(x_l^o,y_l^o)\}$:
\begin{align}
    D_l[x][y] &= \underset{o}{\min}\{\sqrt{(x-x_l^o)^2+(y-y_l^o)^2}\},
\end{align}
Then the ground truth query map $V_l^*$ is defined as
\begin{equation}
    V_l^*[x][y] =
      \begin{cases}
        1 & \text{if~$D_l[x][y] < s_l$}\\
        0 & \text{if~$D_l[x][y] \geq s_l$}
      \end{cases}.       
\end{equation}

For each level $P_l$, the loss function is defined as following:
\begin{footnotesize}
\begin{equation}
	\begin{aligned}
    \mathcal{L}_{l}(U_l, R_l, V_l) = \mathcal{L}_{FL}(U_l, U_l^{*}) + \mathcal{L}_{r}(R_l, R_{l}^*) + \mathcal{L}_{FL}({V_l}, V_{l}^*)
	\end{aligned}
\end{equation}
\end{footnotesize}
where $U_l$, $R_l$, $V_l$ are the classification output, regressor output and the query score output, and $U_l^{*}$, $R_{l}^*$, and $V_{l}^*$ are their corresponding ground-truth maps; $\mathcal{L}_{FL}$ is the focal loss and $\mathcal{L}_{r}$ is the bounding box regression loss, which is smooth $l_1$ loss~\cite{girshick2015fast} in the original RetinaNet. The overall loss is:
\begin{align}
    \mathcal{L}_{all}= \sum_{l}{\beta_l * \mathcal{L}_l}.
\end{align}
Here we re-balance the loss of each layer by $\beta_l$. The reason is that as we add the higher-resolution features like $P_2$, the distribution of the training samples has significantly changed. The total number of training samples on $P_2$ is even larger than the total number of training sample cross $P_3$ to $P_7$. If we don't reduce the weight of it, the training will be dominated by small objects. Thus, we need to re-balance the loss of different layers to make the model simultaneously learn from all layers.

\subsection{Relationships with Related Work}\label{sec:method-3}
Note that though our method bears some similarities with two-stage object detectors using RPN, they differ in the following aspects: 
1), we only compute classification results in the coarse prediction, while RPN computes both classification and regression.
2), RPN is computed on all levels of full feature maps while the computation of our \qdet\ is sparse and selective.
3), two-stage methods rely on operations like RoIAlign~\cite{he2017mask} or RoIPooling~\cite{girshick2015fast} to align the features with the first stage proposal. Nevertheless, they are not used in our approach since we do not have box output in the coarse prediction. It is worth noting that our proposed method is compatible with the FPN based RPN, so \qdet~can be incorporated into  two-stage detectors to accelerate proposal generation.

Another closely related work is PointRend \cite{kirillov2020pointrend}, which computes high-resolution segmentation maps using very few adaptive selected points. The main differences between our \qdet~and PointRend are: 1) how the queries are generated and 2) how sparse computation is applied. For the first difference, PointRend selects the most uncertain regions based on the predicted score at each location, while we directly add an auxiliary loss as supervision. Our experiments show this simple method can generate high recall predictions and improve the final performance. As for the second, PointRend uses a multi-layer perceptron for per-pixel classification. It only requires the features from a single location in high-resolution feature maps, thus can be easily batched for high efficiency. On the other hand, as object detection requires more context information for accurate prediction, we use sparse convolution with $3 \times 3$ kernels.

\section{Experiments}
We conduct quantitative experiments on two object detection datasets: COCO~\cite{lin2014microsoft} and VisDrone~\cite{zhu2018visdrone}. COCO is the most widely used dataset for general object detection; VisDrone is a dataset specialized to drone-shot image detection, in which small objects dominate the scale distribution. 

\subsection{Implementation Details}
We implement our approach based on PyTorch~\cite{paszke2019pytorch} and the Detectron2 toolkit~\cite{wu2019detectron2}. All models are trained on 8 NVIDIA 2080Ti GPUs. For COCO, we follow the common training practices: We adopt the standard $1 \times$ schedule and the default data augmentation in Detectron2. Batch size is set to 16 with the initial learning rate of 0.01. The weights $\beta_l$ used to re-balance the loss between different layers are set to linearly growing from $1$ to $3$ across $P_2$ to $P_7$. For VisDrone, following ~\cite{liu2020hrdnet}, we equally split one image into four non-overlapping patches and process them independently during training. We train the network for 50k iterations with an initial learning rate of 0.01, and decay the learning rate by 10 at 30k and 40k iteration. The re-balance weights $\beta_l$ are set to linearly growing from $1$ to $2.6$. For both datasets, we freeze all the batch normalization (BN) layers in the backbone network during training and we did not add BN layers in the detection heads. Mixed precision training~\cite{micikevicius2017mixed} is used in all experiments to save GPU memory. 
The query threshold $\sigma$ is set to 0.15 and we start query from $P_4$. Without specified description, our method is constructed on RetinaNet with ResNet-50 backbone.

\subsection{Effectiveness of Our Approach}
\begin{table}[t]
  \begin{center}
  \resizebox{\columnwidth}{!}{
    \begin{tabular}{c|c|ccc|ccc|c}
      \hline
      Method & \csqAbrr & AP & AP$_{50}$ & AP$_{75}$ & AP$_{S}$ & AP$_{M}$ & AP$_{L}$ & FPS \\
      \hline
      RetinaNet & - & 37.46 & 56.90 & 39.94 & 22.64 & 41.48 & 48.04 &13.60 \\
      RetinaNet (3x) & - & 38.76 & 58.27 & 41.24 & 22.89 & 42.53 & 50.02 & 13.83 \\
      QueryDet & $\times$ &38.53 & 59.11 & 41.12 & 24.64 & 41.97 & 49.53 &4.85 \\
      QueryDet & \checkmark &38.36 & 58.78 & 40.99 & 24.33 & 41.97 & 49.53 & 14.88 \\
      QueryDet (3x) & $\times$ &39.47 &59.93  &42.11  &25.24  &42.37  &51.12  &4.89  \\
      QueryDet (3x) & \checkmark &39.34 &59.69  &41.98  &24.91  &42.38  &51.12  &15.94  \\
      \hline
    \end{tabular}}
    \end{center}
  \caption{Comparison of accuracy (AP) and speed (FPS) of our \qdet~and the baseline RetinaNet on COCO \textit{mini-val} set.}
\label{tbl-coco-effect}
\end{table}

\begin{table}[t]
  \begin{center}
  \resizebox{\columnwidth}{!}{
    \begin{tabular}{c|c|ccc|cccc|c}
      \hline
      Method & \csqAbrr &AP & AP$_{50}$ & AP$_{75}$  & AR$_{1}$ & AR$_{10}$ & AR$_{100}$ & AR$_{500}$ & FPS  \\
      \hline
      RetinaNet & - &  26.21 & 44.90 & 27.10 & 0.52 & 5.35 & 34.63 & 37.21 & 2.63 \\
      QueryDet & $\times$ &  28.35 & 48.21 & 28.78 & 0.51 & 5.96 & 36.48 & 39.42 & 1.16 \\
      QueryDet & \checkmark &  28.32 & 48.14 & 28.75 & 0.51 & 5.96 & 36.45 & 39.35 & 2.75 \\
      \hline
    \end{tabular}}
    \end{center}
  \caption{Comparison of detection accuracy (AP) and speed (FPS) of our \qdet~and the baseline RetinaNet on VisDrone validation set.}
\label{tbl-visdrone-effect}
\vspace{-0.3cm}
\end{table}

\begin{table*}[t]
  \begin{center}
  %\resizebox{\columnwidth}{!}{
    \begin{tabular}{cccc|ccc|ccc|c}
      \hline
      HR & RB & QH & \csqAbrr &AP & AP$_{50}$ & AP$_{75}$ & AP$_{S}$ & AP$_{M}$ & AP$_{L}$ &FPS\\
      \hline
        &   &   &  & 37.46 & 56.90 & 39.94 & 22.64 & 41.48 & 48.04 & 13.60\\
        \checkmark &   &   &  &36.10 & 56.39 & 38.17 & 21.94 & 39.91 & 45.25 & 4.83\\
        & \checkmark  &   &  &37.66 & 57.57 & 40.37 & 22.03 & 41.86 & 49.10 & 13.60\\
        \checkmark &  \checkmark &   &  &38.11 & 58.48 & 40.85 & 23.06 & 41.53 & 49.36 & 4.83\\
        \checkmark &  \checkmark  & \checkmark  &  &38.53 & 59.11 & 41.12 & 24.64 & 41.97 & 49.53 & 4.85\\
        \checkmark &  \checkmark  & \checkmark  & \checkmark &38.36 & 58.78 & 40.99 & 24.33 & 41.97 & 49.53 & 14.88\\
      \hline
    \end{tabular}%}
    \end{center}
  \caption{Ablation studies on COCO \textit{mini-val} set. {\textbf {HR}} stands for using of high-resolution features; {\textbf {RB}} stands for the loss re-balance between FPN layers; {\textbf {QH}} stands for whether to add QueryHead that provides extra objectiveness supervision.}
\label{tbl-coco-ablation}
\vspace{-0.3cm}
\end{table*}

In Table~\ref{tbl-coco-effect}, we compare the mean average precision (mAP) and average frame per second (FPS) between our methods and the baseline RetinaNet on COCO. The baseline runs at 13.6 FPS, and gets 37.46 overall AP and 22.64 AP$_S$ for small objects, which is slightly higher than the results in the original paper~\cite{lin2017focal}. With the help of high-resolution features, our approach achieves 38.53 AP and 24.64 AP$_S$, improving the AP and AP$_S$ by 1.1 and 2.0. The results reveal the importance of using high-resolution features when detecting small objects. However, incorporating such a high-resolution feature map significantly decrease the inference speed to 4.85 FPS. When adopting our \csqFull, the inference speed is enhanced to 14.88 FPS, becoming even faster than the baseline RetinaNet that does not use the higher-resolution $P_2$, while the performance loss is negligible. Additionally, Figure~\ref{fig:flops} shows how our \csqAbrr~save the computational cost. Compared with the RetinaNet with higher-resolution $P_2$, in which $P_3$ and $P_2$ account for 74\% of the total FLOPs, our \csqAbrr~successfully reduce those costs to around 1\%. The reason is that in \qdet all computations on high-resolution $P_3$ and $P_2$ are carried out on locations around the sparsely distributed small objects. These results sufficiently demonstrate the effectiveness of our method. We also show the results of 3$\times$ training schedule in Table~\ref{tbl-coco-effect}. The stronger baseline does not weaken our improvement but brings more significant acceleration. We owe it to the stronger Query Head as the small object estimation becomes more accurate.

In VisDrone, as illustrated in Table~\ref{tbl-visdrone-effect}, the discoveries are similar, but the results are even more significant. We improve the overall AP by 2.1 and $\text{AP}_{50}$ by 3.2 on this small objects oriented dataset. The inference speed is improved to 2.3$\times$  from 1.16 FPS from 2.75 FPS.

\begin{figure}[!t]
  \includegraphics[width=\columnwidth]{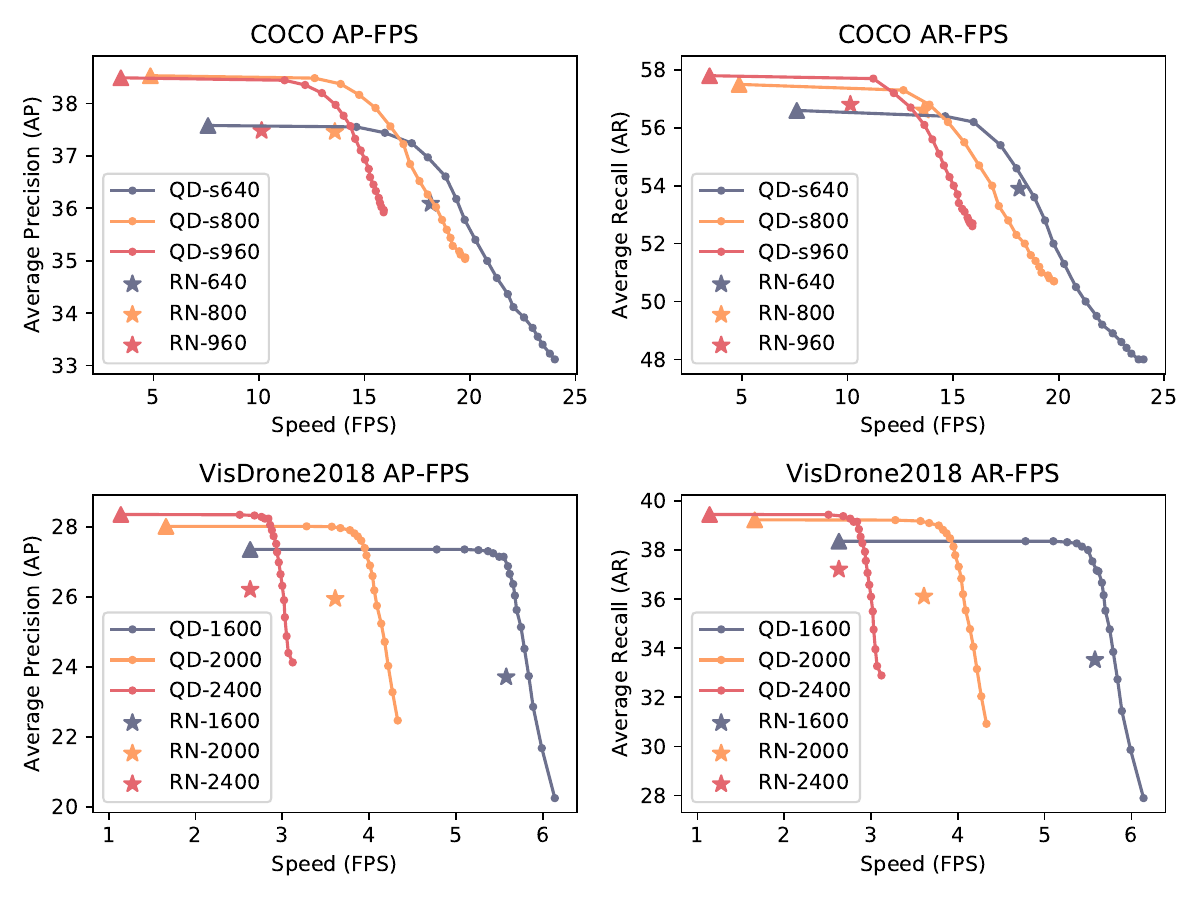}
  \caption{The speed and accuracy (AP and AR) trade-off with input images with different sizes on COCO and VisDrone. The trade-off is controlled by the the query threshold $\sigma$. The leftmost marker (the $\blacktriangle$ marker) of each curve stands for the result when \csqTerm \ is not applied. QD stands for \qdet~and RN stands for RetinaNet. }
  \label{fig:ap-ar-fps}
  \vspace{-0.1cm}
\end{figure}

\begin{table}[t]
  \begin{center}
  \resizebox{\columnwidth}{!}{
    \begin{tabular}{c|ccc|ccc|c}
      \hline
      Start Layer& AP & AP$_{50}$ & AP$_{75}$ & AP$_{S}$ & AP$_{M}$ & AP$_{L}$ & FPS\\
      \hline
        No Query & 38.53 & 59.11 & 41.12 & 24.64 & 41.97 & 49.53 & 4.86 \\
        P$_6$ &  37.91 & 57.98 & 40.51 & 23.18 & 42.02 & 49.53 & 13.42 \\
        P$_5$ &  38.22 & 58.55 & 40.86 & 23.65 & 42.00 & 49.53 & 13.92 \\
        P$_4$ &  38.36 & 58.78 & 40.99 & 24.33 & 41.97 & 49.53 & 14.88 \\
        P$_3$ &  38.45 & 58.94 & 41.07 & 24.50 & 41.93 & 49.52 & 11.51 \\
      \hline
    \end{tabular}}
  \end{center}
  \caption{Investigation of the best starting layer of our \csqAbrr \ on MS-COCO \textit{mini-val} set.}
\label{tbl-coco-start}
\vspace{-0.3cm}
\end{table}

\subsection{Ablation Studies}
We conduct ablation studies on COCO \textit{mini-val} set to analyze how each component affects the detection accuracy and speed in Table~\ref{tbl-coco-ablation}. Our retrained RetinaNet achieves 37.46 AP. When we add the high-resolution $P_2$, the AP dramatically drops by 1.34. As we discussed in Section~\ref{sec:method-2}, this problem is caused by the distribution shift in the training samples after adding $P_2$. Then we re-balance the loss of those layers. The result is improved to 38.11, mostly addressing this problem. Interestingly, the re-balancing strategy only gives us a minor AP enhancement (0.2) when adopting on the original baseline, suggesting that the loss re-balance is more critical in the high-resolution scenario. Then we add our Query Head into the network, through which we get a further performance gain of 0.42 AP and 1.58 AP$_S$, pushing the total AP and AP$_S$ to 38.53 and 24.64, verifying the effectiveness of the extra objectiveness supervision. Finally, with \csqAbrr, the detection speed is largely improved to 14.88 FPS from 4.85 FPS, and the 0.17 loss in detection AP is negligible.  

\subsection{Discussions}

\noindent \textbf{Influence of the Query Threshold.}
Here we investigate the accuracy-speed trade-off in our \csqTerm. We measure the detection accuracy (AP) and detection speed (FPS) under different query thresholds $\sigma$ whose role is to determine if a grid (low-resolution feature location) in the input image contains small objects. Intuitively, increasing this threshold will decrease the recall of small objects but accelerate the inference since fewer locations are considered. The accuracy-speed trade-off with different input sizes are presented in Figure~\ref{fig:ap-ar-fps}. We increase $\sigma$ by 0.05 sequentially for adjacent data markers in one curve, and the leftmost marker denotes the performance when \csqAbrr~is not applied. We observe that even a very low threshold (0.05) can bring us a massive speed improvement. This observation validates the effectiveness of our approach. Another observation is about the gap between the AP upper bound and lower bound of different input resolutions. This gap is small for large size images, but huge for small size images, which indicates that for higher-resolution input our \csqAbrr~can guarantee a good AP lower bound even if the query threshold is set to high.
\newline

\begin{table}[t]
  \begin{center}
  \resizebox{\columnwidth}{!}{
    \begin{tabular}{c|ccc|ccc|c}
      \hline
      Query Method& AP & AP$_{50}$ & AP$_{75}$ & AP$_{S}$ & AP$_{M}$ & AP$_{L}$ & FPS\\
      \hline
        No Query & 38.53 & 59.11 & 41.12 & 24.64 & 41.97 & 49.53 & 4.86 \\
        CQ &  38.31 & 58.73 & 40.98 & 24.25 & 41.98 & 49.53 & 10.49 \\
        CCQ &  38.32 & 58.75 & 40.98 & 24.26 & 41.98 & 49.53 & 8.73  \\
        \csqAbrr \  (ours) & 38.36 & 58.78 & 40.99 & 24.33 & 41.97 & 49.53 & 14.88 \\
      \hline
    \end{tabular}}
  \end{center}
  \caption{Comparison of different query methods on COCO \textit{mini-val} set. We compare our \csqAbrr~and Crop Query (CQ) and Complete Convolution Query (CCQ).}
\label{tbl-coco-query-method}
\end{table}

\begin{table}[t]
  \begin{center}
  \resizebox{\columnwidth}{!}{
    \begin{tabular}{c|ccc|ccc|c}
      \hline
      Context & AP & AP$_{50}$ & AP$_{75}$ & AP$_{S}$ & AP$_{M}$ & AP$_{L}$ & FPS\\
      \hline
        No Query &  38.53 & 59.11 & 41.12 & 24.64 & 41.97 & 49.53 & 4.86 \\
        1x1  &  38.25 & 58.60 & 40.87 & 23.88 & 41.97 & 49.53 & 14.09 \\
        3x3 &  38.30 & 58.66 & 40.94 & 24.14 & 41.97 & 49.53 & 14.06 \\
        5x5 &  38.36 & 58.72 & 40.98 & 24.18 & 41.97 & 49.53 & 14.00 \\
        7x7 &  38.37 & 58.73 & 40.98 & 24.30 & 41.97 & 49.53 & 13.77 \\
        9x9 &  38.37 & 58.73 & 40.98 & 24.30 & 41.97 & 49.53 & 13.42 \\
        11x11 &  38.38 & 58.755 & 40.99 & 24.33 & 41.97 & 49.53 & 13.11 \\
      \hline
    \end{tabular}}
  \end{center}
  \caption{Comparison of detection AP and speed when using different amount of context information on MS-COCO \textit{mini-val} set. The context is defined as a patch with various size around the queried position.}
\label{tbl-coco-context}
\end{table}

\begin{table}[t]
    \begin{center}
    \resizebox{\columnwidth}{!}{
      \begin{tabular}{c|c|c|ccc|ccc|c}
        \hline
        Backbone &Model &\csqAbrr &AP &AP$_{50}$ &AP$_{75}$ &AP$_{S}$ &AP$_{M}$ &AP$_{L}$ & FPS \\
        \hline
        \multirow{3}{*}{MobileNet V2} &RN & - & 26.72 & 43.17 & 28.17 & 15.27 & 29.28 & 34.51 &17.75 \\
                 &QD & $\times$ & 29.16 & 46.20 & 30.95 & 16.14 & 31.26 & 38.66 &5.31 \\
                 &QD & \checkmark & 28.94 & 45.79 & 30.71 & 15.74 & 31.26 & 38.66 & 21.66 \\
        \hline
        \multirow{3}{*}{ShuffleNet V2} &RN & - & 23.04 & 38.32 & 23.75 & 12.01 & 25.50 & 35.16 & 17.45 \\
                 &QD & $\times$ & 26.07 & 42.34 & 27.30 & 13.20 & 28.03 & 36.23 & 5.26 \\
                 &QD & \checkmark & 25.85 & 41.96 & 27.08 & 12.81 & 28.05 & 36.23 & 20.02 \\
        \hline
      \end{tabular}}
      \end{center}
    \caption{Results on different backbone networks. {\textbf{RN}} and {\textbf{QD}} stand for RetinaNet and QueryDet, respectively.}
  \label{tbl-coco-backbone}
  \end{table}

\noindent \textbf{Which layer to start query?}
In our \csqTerm, we need to decide the starting layer, above which we run conventional convolutions to get the detection results for large objects. The reason we do not start our \csqAbrr~from the lowest resolution layer are in two folds: 1) The normal convolution operation is very fast for the low-resolution features, thus the time saved by \csqAbrr~cannot compensate the time needed to construct the sparse feature map; 2) It is hard to distinguish small objects on feature maps with very low resolution. The results are presented in Table~\ref{tbl-coco-start}. We find that the layer that gets the highest inference speed is $P_4$, which validates that querying from very high-level layers such as $P_5$ and $P_6$ would cause loss of speed. We observe that the AP loss gradually increases as the starting layer becomes higher, suggesting the difficulty for the network to find small objects in very low-resolution layers. \\ 

\begin{table}[t]
    \begin{center}
    \resizebox{\columnwidth}{!}{
      \begin{tabular}{c|c|ccc|ccc|c}
        \hline
        ~ & \csqAbrr &AP & AP$_{50}$ & AP$_{75}$ & AP$_s$ & AP$_m$ & AP$_l$ & FPS\\
        \hline
        FCOS & - & 38.37 & 57.63 & 41.03 & 22.34 & 41.95 & 48.96 & 17.06 \\
        QueryDet (FCOS) & $\times$ & 40.05 & 58.69 & 43.46 & 25.52 & 43.43 & 50.69 & 7.92\\
        QueryDet (FCOS) & \checkmark & 39.49 &  57.97 & 42.82 & 24.81 & 43.45 & 50.69 &14.40\\
        \hline
      \end{tabular}}
      \end{center}
      \vspace{-0.3cm}
    \caption{Performance and speed of our QueryDet (FCOS) and its baseline model on COCO \textit{mini-val} set.}
  \label{tbl-fcos}
  \vspace{-0.4cm}
  \end{table}
\begin{table}[t]
\begin{center}
\resizebox{\columnwidth}{!}{
          \begin{tabular}{c|ccc|ccc|c}
          \hline
          \csqAbrr &AP & AP$_{50}$ & AP$_{75}$ & AP$_s$ & AP$_m$ & AP$_l$ & FPS\\
          \hline
          $\times$ & 38.47 & 59.44 & 41.73 & 22.98 & 41.90 & 49.55 & 17.57 \\
          $\checkmark$ & 38.20 & 58.88 & 41.50 & 22.23 & 41.91 & 49.55 & 19.03\\
          \hline
          \end{tabular}}
          \end{center}
\caption{Performance and speed of using our CSQ in Faster R-CNN on COCO \textit{mini-val} set.}
\label{tbl-faster}
\end{table}

\noindent \textbf{What is the best way to use queries?}
We demonstrate the high efficiency of our \csqTerm. We propose two alternative query operations for comparison. The first Crop Query (CQ), in which the corresponding regions indicated by queries are cropped from the high-resolution features for subsequent computations. Note this type of query is similar to the AutoFocus~\cite{najibi2019autofocus} approach. Another one is Complete Convolution Query (CCQ) where we use regular convolutions to compute the full feature map for each layer, but only extract results from queried positions for post-processing. For CQ, we crop a $11 \times 11$ patch from the feature map, which is chosen to fit the receptive field of the five $3 \times 3$ consecutive convolutions in the detection heads. We present the results in Table~\ref{tbl-coco-query-method}. Generally speaking, all three methods can successfully accelerate the inference with negligible AP loss. Among them, our \csqAbrr~can achieve the fastest inference speed. \\

\noindent \textbf{How much context do we need?}
To apply our \csqAbrr, we need to construct a sparse feature map where only the positions of small objects are activated. We also need to activate the context area around the small objects to avoid decreasing accuracy. However, in practice, we found that too much context cannot improve the detection AP but only slow down the detection speed; on the other hand, too little context would severely decrease the detection AP. In this section, we explore how much context do we need to balance the speed-accuracy trade-off. Here, the context is defined as a patch with various sizes around the queried position, where our sparse detection head would also process the features within the patch. The result is reported in Table~\ref{tbl-coco-context}. From it we conclude that a 5x5 patch can brings us enough context to detect a small object. Although more context brings a small AP improvement, the accelerating effect of our \csqAbrr~is negatively affected, while fewer context cannot grantee a high detection AP. \\

\begin{figure*}[t]
\begin{center}
    \includegraphics[width=0.9\linewidth]{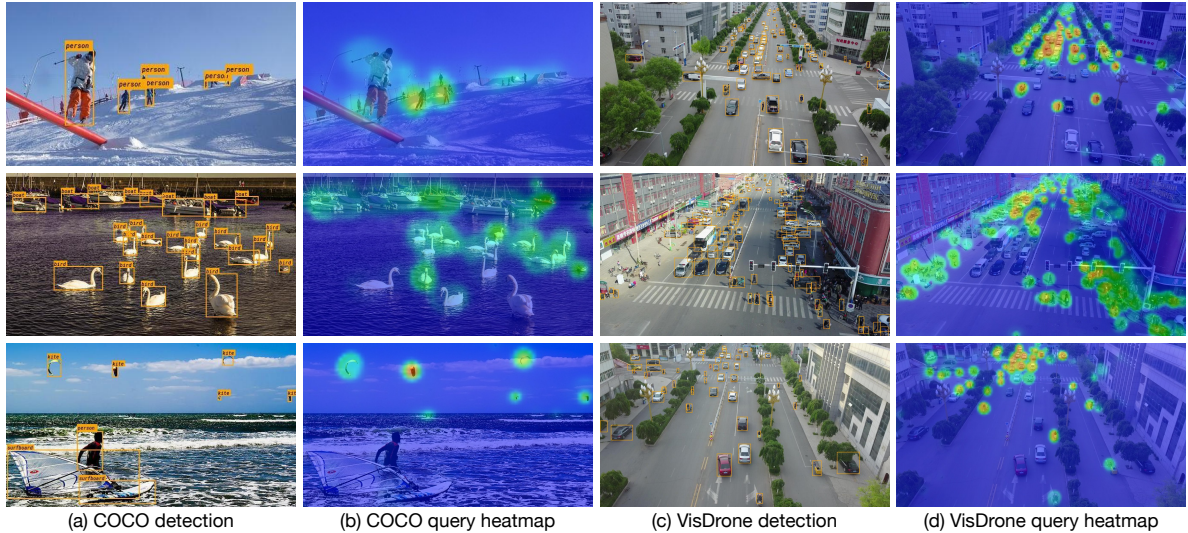}
\vspace{-0.2cm}
    \caption{Visualization of the detection results and the query heatmap for small objects of our \qdet \ on MS-COCO and VisDrone2018 datasets. We remove class labels for VisDrone2018 to better distinguish the small bounding boxes.}
\label{fig:Visualization}
\end{center}
\vspace{-0.5cm}
\end{figure*}

\noindent \textbf{Results on Light-weight Backbones.}
As we claim in Section~\ref{sec:intro}, our method can be incorporated with light-weighted backbones to gain more speed improvement. Also, as our \csqAbrr~aims to accelerate the computation in the detection head, so the overall acceleration is more obvious when using such backbones, because the inference time for backbone network becomes less. We report the results with different light-weight backbones in Table~\ref{tbl-coco-backbone}.Specifically, the speed is on average improved to 4.1$\times$ for high-resolution detection with MobileNet V2~\cite{sandler2018mobilenetv2} and 3.8$\times$ with ShuffleNet V2~\cite{shufflenet}, which validates that our approach is ready to deploy on edge devices for real-time applications such as autonomous driving vehicles for effective small object detections. \\

\noindent \textbf{Results on Anchor-Free Detectors.}
QueryDet can be applied to any FPN based detector to accelerate high-resolution detection. Thus, we apply QueryDet on FCOS, a state-of-the-art anchor-free detector, and report the COCO results in Table~\ref{tbl-fcos}. It can be concluded that \qdet~improves APs with the help of high-resolution features, and when Cascade Sparse Query (CSQ) is adopted, the high-resolution speed is improved by 1.8$\times$ on average, validating the universality of the proposed approach. \\

\noindent \textbf{Effectiveness on Two-stage Detectors}
Our CSQ can also be applied to FPN based two-stage detectors to reduce computaion cost in the high-resolution layers in RPN. To verify this claim, we apply CSQ to the Faster R-CNN detector~\cite{ren2015faster}. In our implementation, the inputs to RPN are from P$_2$ to P$_6$ and we start query from P$_4$. We modify the RPN structure to let it have 3 \textit{conv} layers instead of 1 layer in the normal implementation, which is followed by 3 branches for objectiveness classification, bounding box regression and query key computation. The former two branches are trained following common practice~\cite{ren2015faster}, and the query branch is trained by Focal Loss with $\gamma =1.2$ and $\alpha =0.25$. During inference, we set the query threadhold to 0.15. As shown in Table~\ref{tbl-faster}, our Faster R-CNN achieves 38.47 overall AP and 22.98 AP$_S$ with 17.57 FPS. When CSQ is utilized, the inference speed is improved to 19.03 FPS with a minor loss in AP$_s$. The results verify the effectiveness of our approach in accelerating two stage detectors. Note that in two-stage detecotrs our CSQ can not only save time for the dense computaion on in RPN, it can reduced the number of RoIs that are fed into the second stage.

\subsection{Visualization and Failure Cases}
In Figure~\ref{fig:Visualization}, we visualize the detection results and the query heatmaps for small objects on COCO and VisDrone. From the heatmaps, it can be seen that our query head can successfully find the coarse positions of the small objects, enabling our \csqAbrr~to detect them effectively. Additionally, through incorporating high-resolution features, our method can detect small objects very accurately.

We also show two typical failure cases of our approach: 1) Even if the corase position of small objects is correctly extracted by the query head, the detection head may fails to localize them (the second image of VisDrone); 2) Positions of large objects is falsely activated, causing the detection head to process useless positions and hence slowing down the speed (the first image of COCO).

\section{Conclusion}
We propose \qdet~that uses a novel query mechanism \csqFull~to accelerate the inference of feature pyramid-based dense object detectors. \qdet~enables object detectors the ability to detect small objects at low cost and easily deploy, making it practical to deploy them on real-time applications such as autonomous driving. For future work, we plan to extend \qdet~to the more challenging 3D object detection task that takes LiDAR point clouds as input, where the 3D space is generally sparser than 2D image, and computational resources are more intense for the costly 3D convolution operations.

{\small
\bibliographystyle{ieee_fullname}
\bibliography{capital_bib}
}

\end{document}